\documentclass[letterpaper, 10 pt, conference]{ieeeconf}  
\usepackage{amsmath}
\usepackage{booktabs}
\usepackage{algorithm}
\usepackage{algpseudocode}
\usepackage{graphicx}
\usepackage{subfig}
\usepackage{multirow}
\usepackage{color} 

\IEEEoverridecommandlockouts                              

\title{\LARGE \bf
	Towards navigation without precise localization: Weakly supervised learning of goal-directed navigation cost map
}

\author{Huifang Ma, Yue Wang,  Li Tang, Sarath Kodagoda, and Rong Xiong 
	\thanks{Huifang Ma, Yue Wang, Tang Li, and Rong Xiong are with the State Key Laboratory of Industrial Control Technology and Institute of Cyber-Systems and Control, Zhejiang University, Zhejiang, China. Sarath Kodagoda is with the Centre for Autonomous Systems, The University of Technology, Sydney, Australia  Yue Wang is the corresponding author {\tt\small wangyue@iipc.zju.edu.cn}. Rong Xiong is the co-corresponding author.}
}

\begin{document}

	\maketitle
	\thispagestyle{empty}
	\pagestyle{empty}

	\begin{abstract}
		Autonomous navigation based on precise localization has been widely developed in both academic research and practical applications. The high demand for localization accuracy has been essential for safe robot planing and navigation while it  makes the current geometric solutions less robust to environmental changes. Recent research on end-to-end methods handle raw sensory data with forms of navigation instructions and directly output the command for robot control. However, the lack of intermediate semantics makes the system more rigid and unstable for practical use. To explore these issues, this paper proposes an innovate navigation framework based on the GPS-level localization, which takes the raw perception data with publicly accessible navigation maps to produce an intermediate navigation cost map that allows subsequent flexible motion planning. A deterministic conditional adversarial network is adopted in our method to generate visual goal-directed paths under diverse navigation conditions. The adversarial loss avoids the pixel-level annotation and enables a weakly supervised training strategy to implicitly learn both of the traffic semantics in image perceptions and the planning intentions in navigation instructions. The navigation cost map is then rendered from the goal-directed path and the concurrently collected laser data, indicating the way towards the destination. Comprehensive experiments have been conducted with a real vehicle running in our campus and the results have verified the robustness to localization error of the proposed navigation system.
	\end{abstract}

	\section{Introduction}
	Precise localization is widely acknowledged as crucial for safe robot navigation based on a pre-built map. This has been extensively tested in many real systems and proven to deliver satisfactory performance. Many conventional systems rely on the transformation of global reference paths into robot local coordinate. They often depend on landmarks for precise localization to successfully maneuver in environments. The navigation becomes challenging and run rigid in scenarios where there are none or minimal landmarks or features.
	
	It is worth investigating how humans drive in complex environments. They do not use precise localisation yet manage to perform excellent navigation tasks. This is different to conventional robotic navigation tasks where a robot makes navigation decisions purely based on geometry. Humans may rely on GPS like sensors to provide rough location information as an aid in navigation, however they also use other cues like road semantics. This approach becomes very important specially when the road environment is lacking land marks or unique features.
	With the trend of deep learning, the end-to-end navigation from input camera images to control command is considered which has incorporated the environment semantics. The drawback lies in the deficiency of intermediate information which cannot guarantee the safety and appropriateness of navigation decisions. Furthermore, it can be complex for the network to take both obstacles avoidance and motion properties into consideration. 
	
\begin{figure}[!t]
		\centering
		\includegraphics[width=0.49\textwidth]{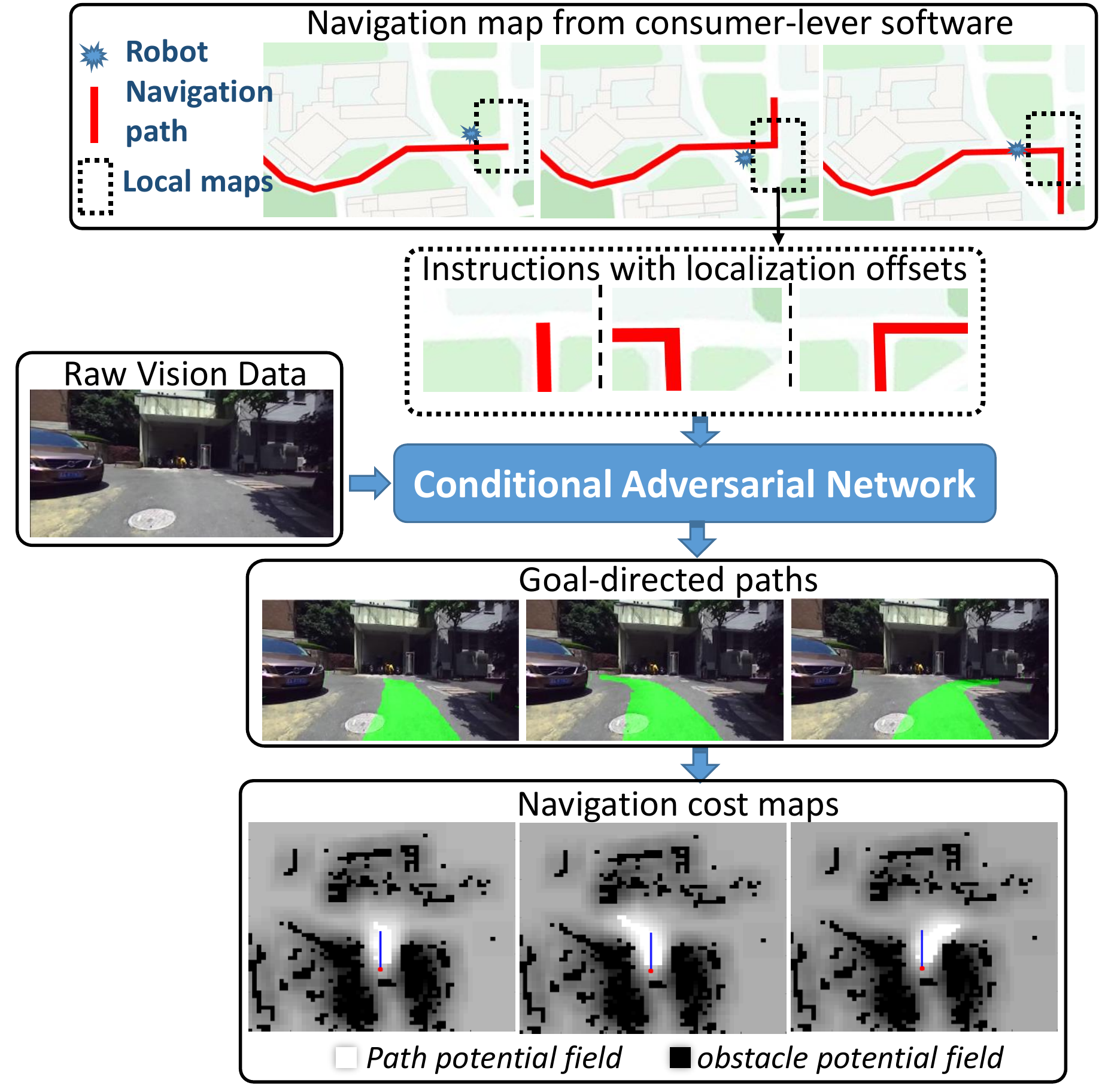}
		\caption{Outline of the proposed navigation framework. The proposed framework imitates the human navigation style which leverages the publicly available navigation map and GPS localization information to get local navigation instructions towards the destination. Vision goal-directed path and top-view navigation cost map are generated without precisely geometric calculation.}
		\label{outline}
	\end{figure}

	Inspired by the human navigation and the difficulties in the learning based navigation system, we set to develop an intermediate learning framework which utilises semantic information entraining to alleviate issues relating to the end-to-end approaches. The outline of the proposed method is shown in Figure \ref{outline}.
	The method firstly predicts the visually goal-directed path given the current image, the approximate GPS localization as well as a reference path, which can be converted to the ground floor path potential field in the local coordinates. The path potential field is an imitation to the rough guidance of the goal in human navigation. Then the obstacles are explicitly detected by the other sensors and represented by another layer of potential, which is fused with the predicted potential of the goal to generate the navigation cost map. The conventional motion planning and control techniques can be employed based on this navigation map to ensure smooth and safe robot actuation. In the proposed framework, the navigation complexity has been reduced by decoupling the states of robot and obstacles in the problem space, which has intrinsically incorporated the road semantics, including traversable regions, free space, obstacle regions, etc. into navigation planning. 
	
	Specifically, the learning model is fed by the row vision image with a 2D reference path provided by consumer lever software. The reference path can have some localization offsets from robot global position. Thus a concise direction semantic from the navigation instruction needs to be extracted and applied to current perception to find a possible path met with extracted road semantics. The challenges are that environment can have potentially infinite moving paths even with a specified instruction, as roads are usually much broader than the vehicle. The predicted path may only need to meet the requirements that it is traversable and heading on the correct direction. As most segmentaion networks have two outcomes per-pixel, we have adopted a deterministic conditional adversarial network to specifically address the problem in a weakly supervised manner which allow path variations to the ground truth.
	\begin{figure*}[!t]
		\centering
		\includegraphics[width=0.95\textwidth]{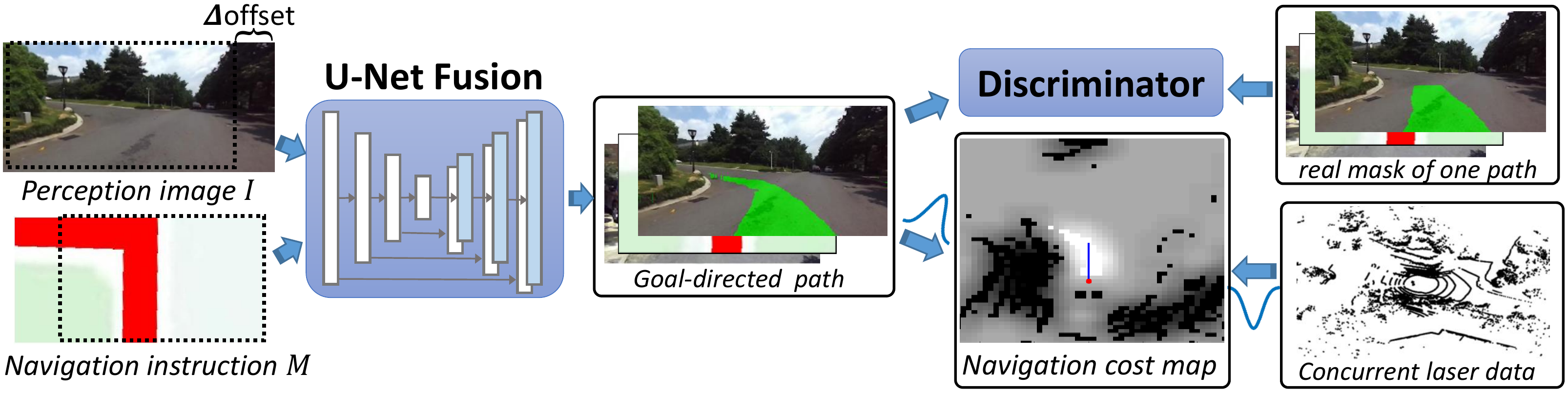}
		\caption{The structure of our model. Raw vision perception is combined with a local navigation instruction and fed into a UNet structure to generate goal-directed path corresponds to the planning intention. Both of the inputs can be randomly cropped to increase the robustness on localization. Then the fake path and one real path are individually combined with the inputs for discrimination. Finally the vision goal-directed path is trasformed to the geometric ground floor and fused with the laser data to get the top-view navigation cost map using a Gaussian kernal.}
		\label{model}
	\end{figure*}
	
	Both the path potential field and the navigation cost map are experimentally validated based on the data collected on a real vehicle. In order to perform comprehensive analyses, an off-line navigation map rendering method is developed to provide instructions with various settings. The result shows the proposed method can give goal-directed paths with a heading direction error less than 10 degree, which is robust to the localization error within 5m. And the cost map has achieved 0.84 and 0.68 probability on prediction of human driver trajectory in the range of 5m and 10m respectively. In conclusion, our contributions can be explicitly stated as follows:
	\begin{itemize}
		\item A learning based navigation system imitating the human navigation is designed, which utilizes an approximate localization given by a low-cost GPS, and yields a potential field of the goal to be integrated with the obstacle for final motion planning and control. The system eliminates precise localization which is required in conventional navigation systems.
		
		\item A deterministic conditional adversarial network is proposed to generate paths under diversely specified navigation instructions. With this network, weakly supervised learning is achieved without human intervention.
		\item A local navigation map rendering method is developed and implemented, which can produce massive navigation instructions for the human-like navigation research.
	\end{itemize}
	
	The remainder of the paper is arranged as follows: Section \ref{related work} introduces the related literatures, Section \ref{method} provides a detailed illustration of our method, experimental results and performance analysis are presented in Section \ref{experimental result}, and a brief conclusion is drawn in Section \ref{conclusion}.
	
\section{Related Work}\label{related work}
Conventional pipeline of robot navigation includes localization, perception and motion planning. The result of localization, i.e. a reference path or a goal in local robot coordinates, and the result of perception, i.e. a grid map with(out) semantics, are fed into motion planning to generate final control command. Localization addresses the geometric transformation calculation, while perception deals with the scene parsing. For localization, both LiDAR based methods and camera based methods are considered. In \cite{wolcott2017robust,wan2018robust,castorena2018ground}, the LiDAR data was firstly converted to a high resolution map by encoding the reflectivity and the height. Then the current scan can was aligned with the map through a similarity criteria to calculate the precise robot pose. Visual localization is extensively investigated in\cite{Furgale2010Visual,Churchill2013Experience,ding2018laser,tang2019topological}. In these works, a visual map is built first through bundle adjustment, in which the robot matches the features observed in the current view to that in the map to estimate its pose. The additional semantic information has also been considered in the work of \cite{li2018stereo,xiang2017darnn}, in order to improve the feature matching confidence. These methods preliminarily focused on improving localization accuracy as well as robustness. Nevertheless, localization actually brings the goal into the robot coordinate to facilitate navigation, i.e. a cost term in planning, instead of the accurate robot pose itself. Therefore, in our method, we set to explore the feasibility of directly learning a navigation cost map induced by a specified goal and to relax the demand of precise localization.

For the traffic scene understanding, one category of methods aim at recognition of traffic participants in the image, such as vehicles, pedestrians and cyclists \cite{redmon2018yolov3,liu2016ssd,chen2018deeplab,luo20173d,li2018scale,yang2016exploit}. Another idea for navigation is to identify the traversable regions, which is more straightforward for robots in off-the-road environments. Wang et al, \cite{wang2017scalable} proposed an online learning mechanism to deal with the appearance change without referring to the massive data. Tang et al, \cite{tang2017one} applies mapping techniques to automatically generate weakly supervised data, which was utilized to supervise the learning of traversable regions. There are also methods focused on holistic scene understanding. In \cite{geiger20143d}, the intersection topology and geometric parameters, as well as the traffic signs and road participants were inferred by a graph model using a set of hand-crafted image features. The work in \cite{chen2015deepdriving} intended to find an intermediate traffic semantics by regression of distances to the lanes as well as other road users. These works have all yielded an intermediate local navigation cost map for robot. While the errors in localization may contribute to plan reference paths through obstacles, leading to unsafe routes. In our work, the goal is to directly learn a cost map with the obstacle information incorporated, rendering the system more robust to localization errors.

Recently, end-to-end method becomes popular in autonomous driving, in which the main argument is that the performance of the intermediate stages in the conventional system architecture may not be aligned with the ultimate goal, namely, the control of the robot. With this idea, Codevilla, et,al.\cite{Codevilla2017End} proposed to directly learn the driving model to compute motion command via conditional imitation learning by considering the repeatability of imitation learning and high-level command input. The work in \cite{gao2017intention} collected control commands from existing local planner (Dynamic Window Approach\cite{fox1997dynamic}) and proposed a two-stage manner to relax the prior information in precise localization, which relies on the path-planning result as the navigation form to learn the supposed motion command based on a residual neural network. The work in \cite{hecker2018end} adopted 360-degree surround-view cameras along with the planned routes information from commercial maps to learn an end-to-end driving model. Their work also utilized GPS signals as well as public map to generate the steering angles and speeds based on a RNN. As reported in their evaluation, it has unavoidably intervened by humans. The work in \cite{amini2018variational} proposed to use a variational network to get a full probability distribution over the possible control commands; however, when combined with specific navigation indicators, they still solve an accurate form of certain angular speed. Compared with this line of methods, our work still focuses on the intermediate stage, while using a learned cost map instead of the precise geometric transformation. We consider the problem space of end-to-end control learning is more complicated than ours, as the motion states of the robot are coupled with the understanding of traffic semantics. Thus it is more difficult for the model to capture semantic relations and is harder to be tested.

	\section{Methods}\label{method}
	We frame the problem of goal-directed path learning as deterministic cGAN (conditional generative adversarial network). The structure of our model is shown in Figure \ref{model}. The environment perception along with a deterministic navigation instruction have been both fed into a UNet for feature extraction. The generated path are combined with the inputs to be processed by the discriminator. Then the goal-directed path is converted to the real world with path potential field with a Gaussian kernel and integrated with the obstacle potential from laser data to produce the navigation cost map. The following sections provide detailed illustration of our method.

	\subsection{Conditional Path Learning} In this section, we describe the conditional adversarial network model used in our approach, which takes raw vision perception images, $I$, and the navigation instruction represented by a 2D local Map, $M$, as input, and outputs visual goal-directed path $P$ towards the goal. The adversarial property can discriminate the prediction with the input condition as a whole, and the deterministic condition from $M$ ensures the generated path is goal-directed.
	
	GANs are generative models that learn a mapping from random noise vector $z$ to output image $y$, $G : z\rightarrow y$\cite{goodfellow2014generative}. Conditional GANs introduce a prior knowledge as observed image $x$ to learn the mapping, $G : \{x,z\}\rightarrow y$. Thus the adversarial network structure can translate random noise into corresponding output which has maintained the original structural attributes of conditional image,  $x$. Inspired by this, we have untilized the deterministic navigation instruction as generation seed to produce corresponding goal-directed path, $G_d : \{I,M\}\rightarrow P$. 
	
	The model structure has basically referred to the work in \cite{isola2017image} with a UNet for generation and a patch-wise loss for discrimination. The UNet\cite{ronneberger2015u} is an encoder-decoder structure with skip connections to consider the low-dimension features, as implied in the Figure \ref{model}. And the objective function can be expressed as the sum of two weighted loss:
	\begin{equation}
	L = arg\,min_{G_d}\,max_D\,L_{cGAN}(G_d,D) + \lambda L_{L1(G_d)}
	\end{equation}
	where $\lambda$ is the weight parameter.
	
	The first item is the standard cGAN objective function:
	\begin{equation}
	\begin{aligned}
	L_{cGAN}(G_d,D) = &E_{I,M,P}[logD(I,M,P)] + \\
	&E_{I,M}[log(1-D(I,M,G_d(I,M)))]
	\end{aligned}
	\end{equation}
	
	and the second item is a patch-wise $L1$ distance\cite{isola2017image} from generated path to the provided real path.
	
	The deterministic cGAN has ensured the generated path to consider both the current structure in the visual perception and the different direction provided in the navigation instruction. It has avoided the pixel-to-pixel match for traversable area segmentation and is trained in a weakly supervised manner to implicitly learn the road semantics. The network can resultantly provide corresponding goal-directed paths under diverse navigation instructions.
	
	\subsection{Navigation Cost Estimation}
	The goal-directed path with specific direction is intuitive for human to follow while lacks geometric information for a robotic vehicle. In order to estimate monocular area in a real world coordinate, we assume that the ground in the near front of the robot can be modeled with a flat plane. Since the method processes perception for the next-step navigation, most of the near fluctuation on ground floor can be omitted. Thus the geometric traversable points has a global height of $0$, and the geometric area in the robot coordinate can be obtained with the camera calibration parameters. Besides, we have incorporated the concurrently collected laser data and defined an obstacle height range to cast the related laser points onto the ground floor. The top-view grid map can therefore be established. After applying a positive Gaussian kernel to the path grids and a negative Gaussian kernel to the obstacle grids, we can estimate a complete navigation cost map in the robot coordinate, which can be used for the following motion planning. 
	
	\subsection{Weakly-supervised label generation}\label{labelgeneration}
	\subsubsection{Navigation instruction rendering}
\begin{figure}[!h]
		\centering
		\subfloat[Navigation route]{\includegraphics[width=0.22\textwidth]{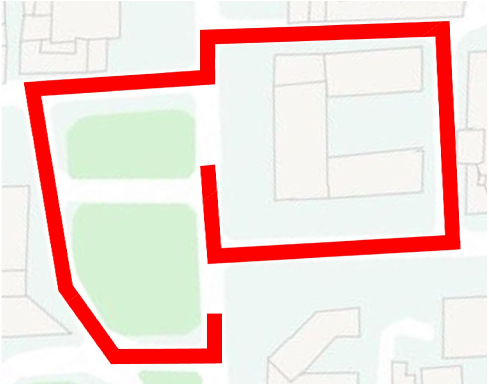}}
		\hspace{0.005\textwidth}
		\subfloat[Discretized route points]{\includegraphics[width=0.22\textwidth]{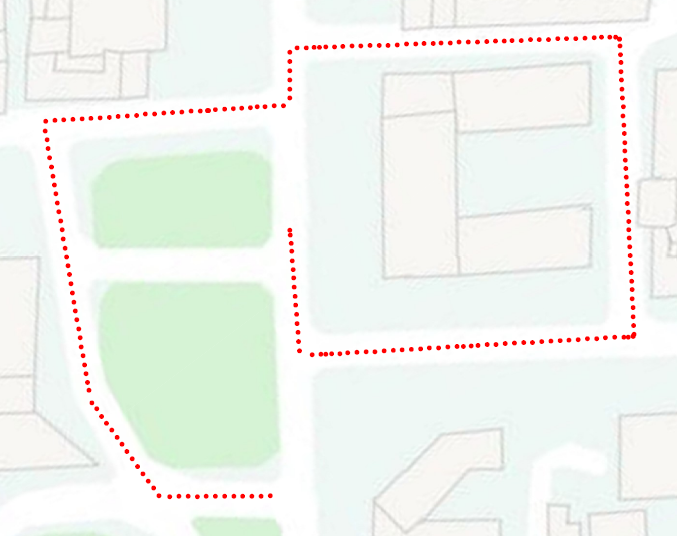}}	
		\\
		\subfloat[Real robot poses]{\includegraphics[width=0.22\textwidth]{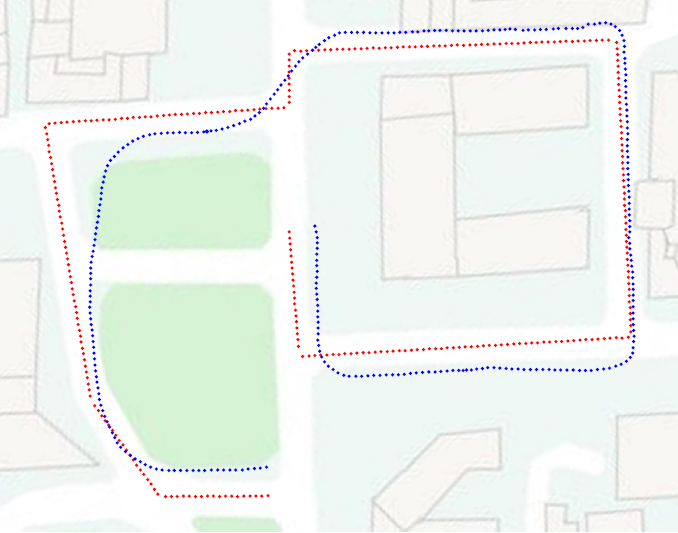}}	
		\hspace{0.005\textwidth}
		\subfloat[Alignment with center positions]{\includegraphics[width=0.22\textwidth]{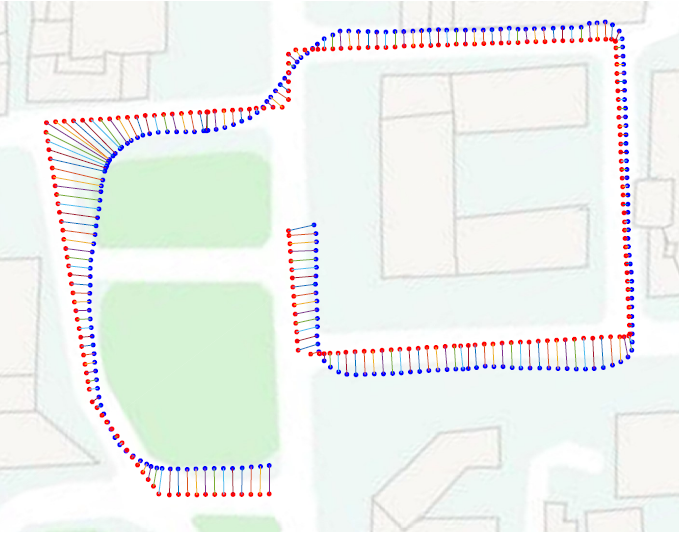}}
		\caption{Local navigation instruction generation.}
		\label{map rendering}	
	\end{figure}

	In relation to a certain place, there can be many navigation instructions due to complexity of the road scenarios. 
	Further, the localization errors need to be carefully considered. 
	This can lead to substantial workload for online data collection and rendering. Therefore, we have developed a local navigation view rendering method, which can facilitate cropping expected local view in different settings. The procedure is shown in Figure \ref{map rendering}. 
	
	Firstly, we have gathered the plan view of the complete 2D map of the data collection route from the publically available Baidu map website. The data collection route has been previously defined and we can manually draw the navigation route on the map (Figure \ref{map rendering}(a)), denoted as $R$. The heading direction is labeled on each near-straight road segments, which is used later in the local map cropping with a correct direction. The next step is to estimate an associated position on $R$ for each of our collected data pose. To achieve this, we have firstly discretized $R$ to route points $R_d$, as shown with red dots in Figure \ref{map rendering}(b). It is obtained by plotting the route center line with red dotted lines and clustering pixels inside each dot to a get the pixel-level single point. Then the geometric robot poses can be projected to the map  (blue line in Figure \ref{map rendering}(c)), denoted as $T_r$, were spatially aligned with the discrete route points $R_d$ by adopting a geometric distance function in the DTW (dynamic time warping) algorithm:
	
	\begin{equation}
	DTW(T_r,R_d) = min\,\frac{1}{K}\sqrt{\sum_{k=1}^{K}w_k}
	\end{equation}
	where $K$ is the warpping length, and $w_k=(i,j)_k$ defines the mapping from $T_r$ to $R_d$. 
	
	The optimiztion objective is:
	\begin{equation}
	\begin{aligned}
	\gamma(i,j) = &d(T_r(i),R_d(j)) +\\ &min\{\gamma(i-1,j-1),\gamma(i-1,j),\gamma(i,j-1)\}
	\end{aligned}
	\end{equation}
	where $\gamma$ is the accumulated series distance and $d$ is the specified distance function. We have defined $d$ as the euclidean distance from the projected robot pose to the center route point:
	\begin{equation}
	d(T_r(i),R_d(j)) = ||(T_r(i)),R_d(j)||_2
	\end{equation}
	
	The aligned result is shown in Figure \ref{map rendering}(d). By considering the heading directions, we can crop the desired local navigation views under different experiment settings.
	
	\subsubsection{Real path collection}
	To collect the real path mask, we have adopted an automatic label annotation method with the assumption that area once covered by the moving vehicle is certainly traversable. The robot trajectory can be transformed onto the camera image for traversable path generation. The other area on the image are labeled with \textit{obstacle} and \textit{unknown} utilizing the projection of current laser data. The specific implementation of real path generation process can be found in our previous work\cite{tang2017one}. The training route in our experiment only covers a single direction for each cross while we can achieve generalization to diverse navigation cases in the test route.

	\subsection{Training with data augmentation}
	Since the weakly-supervised label generation provides only one path with a single navigation instruction, we have introduced a random cropping strategy to reinforce the road structure learning while training as implied in the Figure \ref{model}. The input $I$ and $M$ have been horizontally cropped with a different offset value which is constrained to a limited maximum rate. The randomness helps the model to incorporate more structural semantics of the scene, which avoids the pixel-to-pixel correlation with a fixed view and increased the robustness of the model for both new scenarios and situations with higher localization errors. It is to be noted that we have adopted a horizontal offset to cater for far away junctions. 
	
	For the parameter optimization, we have followed the common procedure of one gradient descent on $D$ and then one step on $G$. The network is trained with stochastic gradient descent(SGD) with a learning rate of 0.0002, and momentum parameters $\beta_1 = 0.5$, $\beta_2 = 0.999$. We have trained each model around 200 epochs with a batch size of 12. At inference time, we ran the generator net in exactly
	the same manner as during the training phase.
	
	\section{Experiments}\label{experimental result}
	This section reports the experiment results of our proposed weakly supervised learning framework, including experiment set up, vision path prediction performance, and real trajectory prediction test based on the navigation cost map. 
	
	\subsection{Set up}
	\subsubsection{Data sets}
\begin{figure}[!h]
		\centering
		\includegraphics[width=0.47\textwidth]{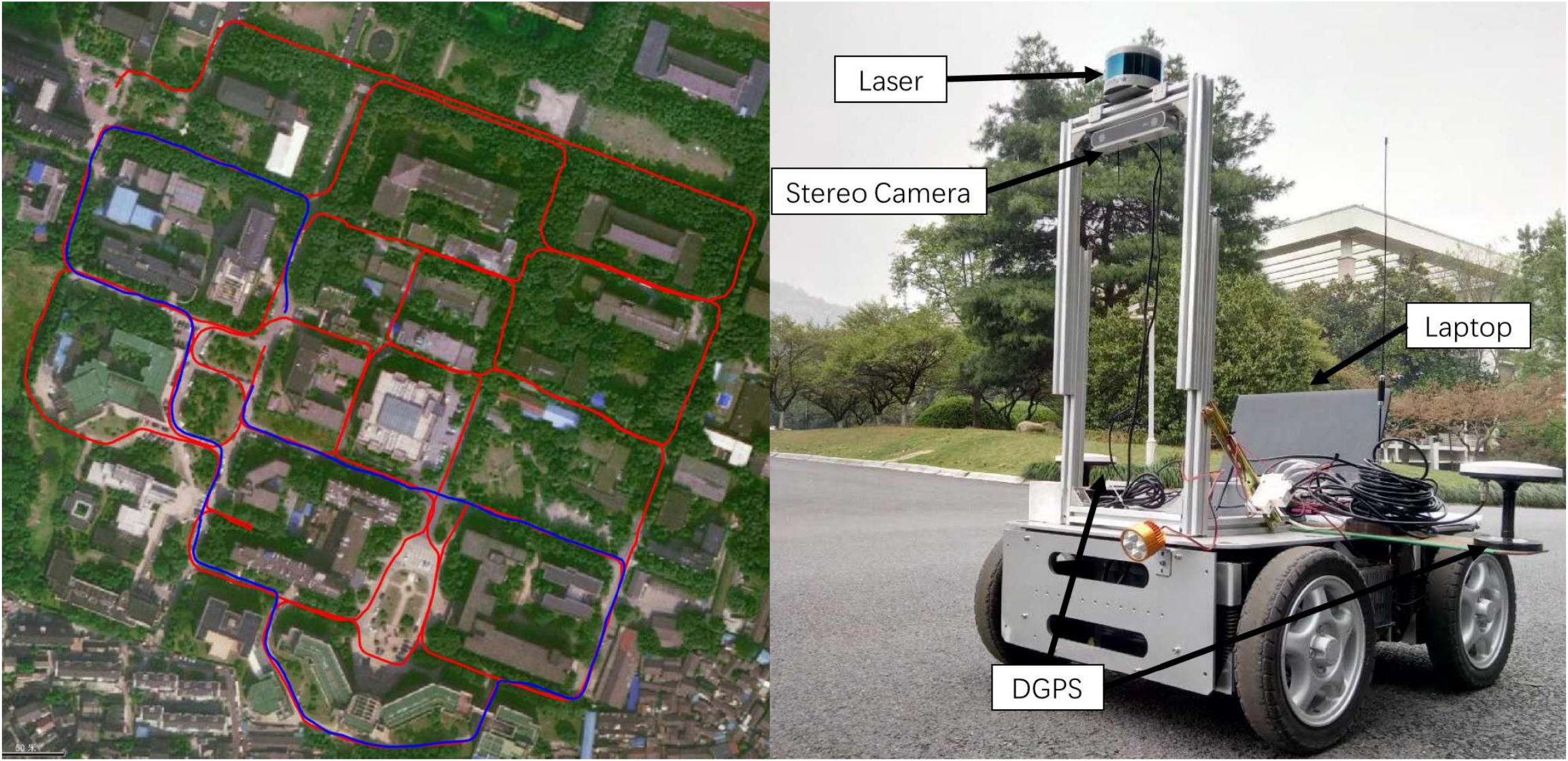}
		\caption{Data collection route and experiment vehicle.}
		\label{hardware}
	\end{figure}

	We test our method based on the perception data from a real vehicle running in our campus. The data collection route is shown in the left of Figure\ref{hardware}. Blue line shows the training route with a length of 1.2km and red line shows the test route with a length of 4.8km, and the overlap sections represents bi-directional driving. The vehicle used for data collection is shown in the right side of Figure\ref{hardware}, which is a four-wheeled mobile robot equipped with a ZED stereo Camera, a Velodyne VLP-16 laser scanner and a D-GPS. Only images from the left camera of ZED are used with a resolution of 314$\times$648. The training data contains 21 groups of perception data, with each one has $\sim$10000 frames of observation. The test data contains one group of $\sim$36000 frames of observation.  
	\subsubsection{Evaluation Metrics}
	We employ three criterion for visual goal-directed path evaluation: the IOU (intersection of union) between predicted path area with provided real path mask, the coverage of predicted central line in the real mask  $cover\_rate$, and the difference of yaw angle $\Delta{yaw}$ under the visual path instruction. The IOU is widely used in vision community for pixel-level evaluation and can imply the similarity with real path and provide the scale evaluation. The $cover\_rate$ reflects the consistency of main direction with the real path. And the $\Delta{yaw}$ shows the difference of driving direction under visual path instruction, which is estimated by projecting the path central line to the ground plane and calculating the yaw angle difference with that of the real path.
	
	To test the performance of navigation cost map, we have estimated the NLL (Negative Log-Likelihood) of the test demonstration trajectory in a distance of 5m and 10m, which is normally required for the motion planning. Moreover, many qualitative results are provided to better illustrate our method. 
	
	\subsection{Goal-directed path generation}
	We have tested our method with three models. The first, $gan\_basic$ model, which is the pix2pix model with multi-input. The other two models are $gan\_rand\_navi$ and $gan\_rand\_per$, which have considered the random offset illustrated in Section\ref{method} for better generalization. The $gan\_rand\_navi$ has adopted random image offset for navigation instruction only, and the $gan\_rand\_per$ has introduced offset for both inputs of perception and navigation. The navigation instruction in this part is provided with a central route view under the assumption of precise localization and the result is presented in Table \ref{path generation}.
	\begin{table}[!h]
		\caption{P{\scriptsize ERFORMANCE} {\scriptsize ON} V{\scriptsize ISION}  P{\scriptsize ATH}  G{\scriptsize ENERATION}}
		\centering
		\begin{tabular}{lccc}
			\toprule[0.025cm]
			Model &IOU/\%&$cover\_rate/$\%&$\Delta{yaw}/deg$\\	
			\cmidrule{1-4}
			\textit{basic} & 60.57&98.85&8.91\\		
			\cmidrule{1-4}
			\textit{$rand\_navi$} & 61.29&98.72&10.75\\
			\cmidrule{1-4}	
			\textit{$rand\_per$} & 54.56&96.31&8.94\\
			\bottomrule[0.025cm]
		\end{tabular}
		\label{path generation}
	\end{table}
	
\begin{figure}[!h]
		\centering
		\subfloat[Traverse different turns. ]{\includegraphics[width=0.48\textwidth]{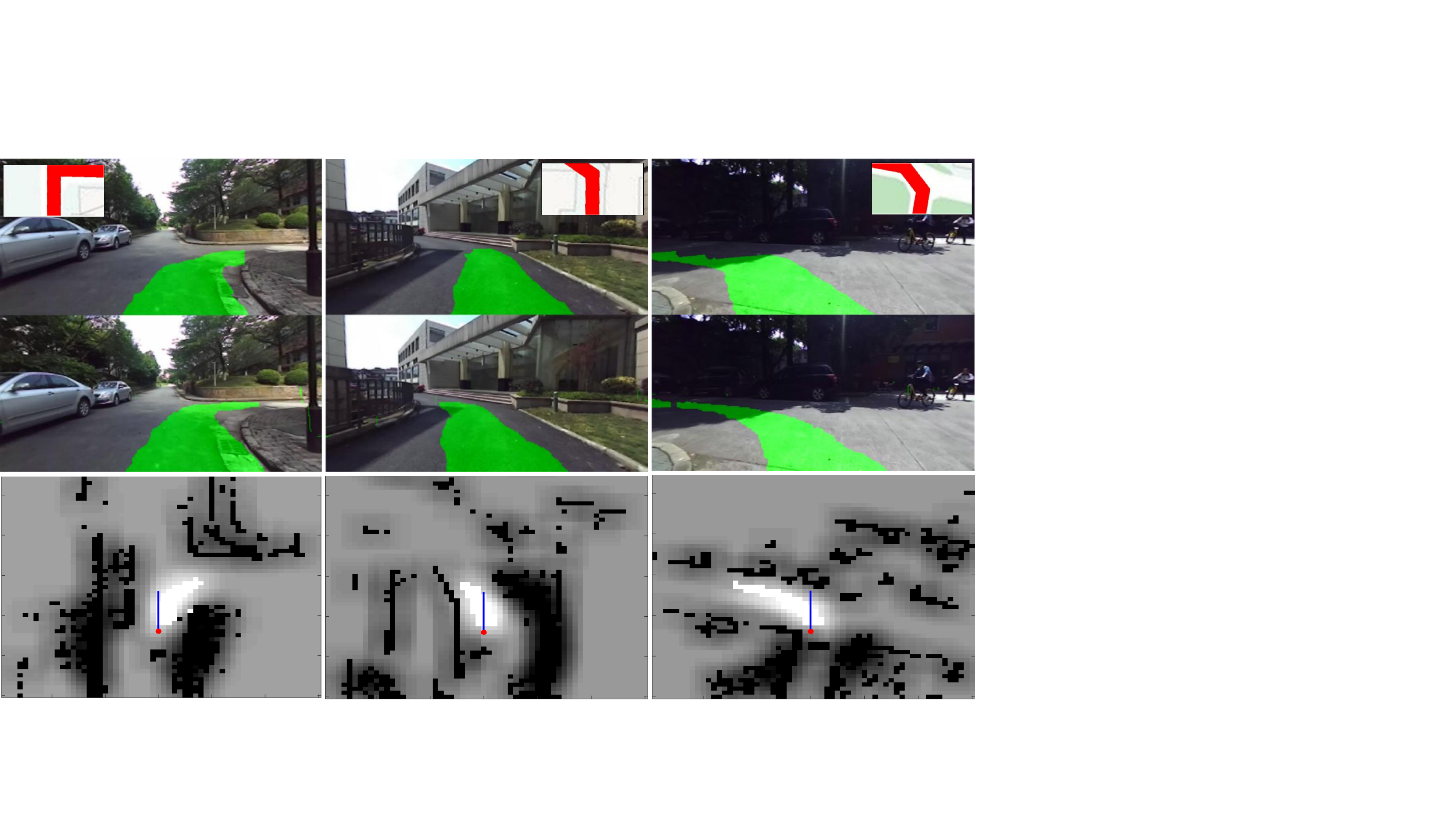}}\\
		\subfloat[Avoid the obstacles.]{	\includegraphics[width=0.48\textwidth]{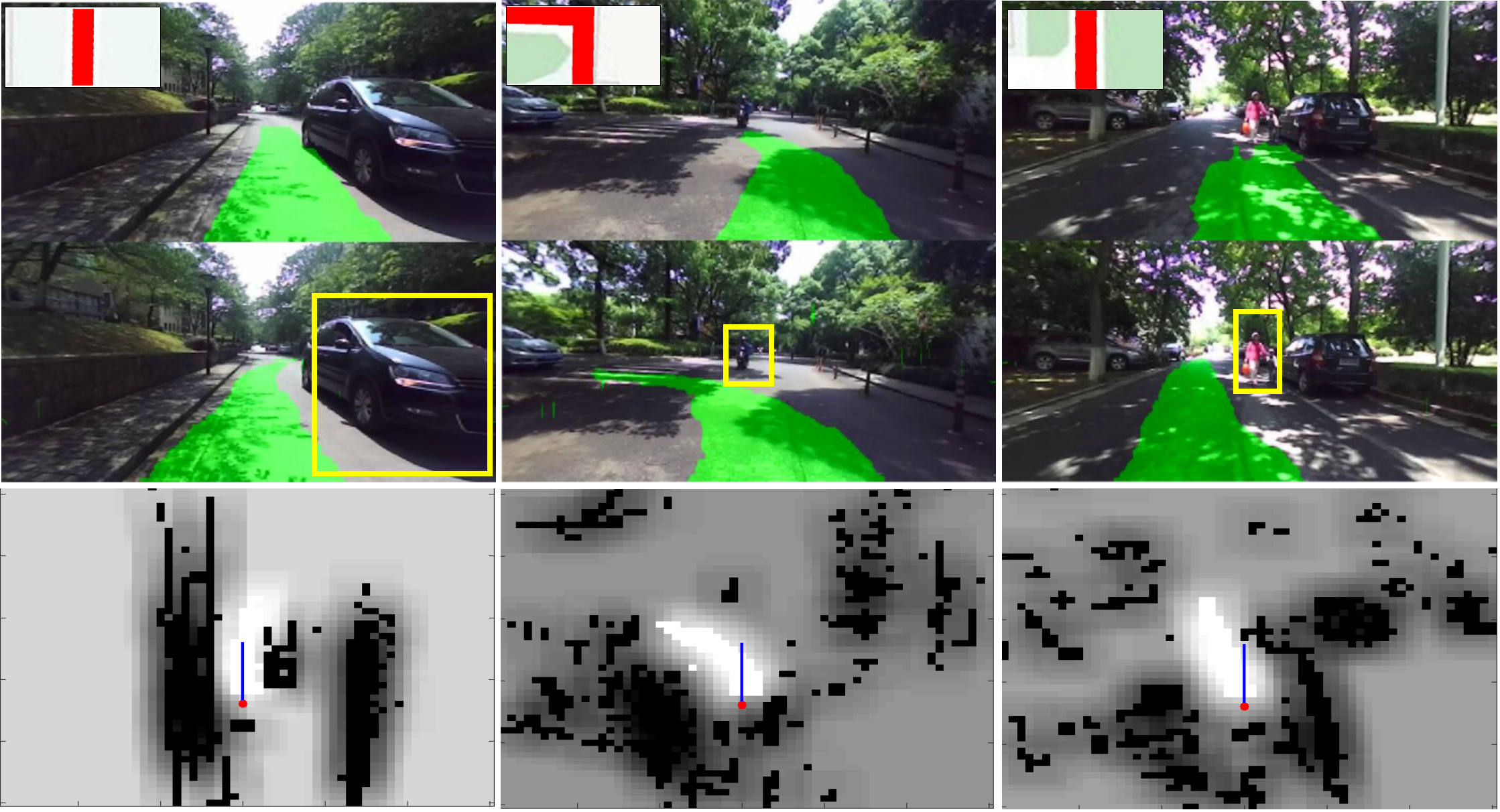}}	\caption{Results of the proposed framework. From top to bottom: the vision ground truth mask with specified navigation structure, the predicted vision path, the generated potential map. The red point on the potential map indicates robot's transformed global position and the blue line implies the robot local coordinates.}
		\label{result_turn}
	\end{figure}

	As shown in the table, the coverage rates of center line have all exceeded 95\%, with more than 50\% IOU with the real path collected. This implies the predicted path is spatially close to the real path. The geometric error of yaw angle is approximately 9 degrees, and the basic model has behaved relatively best for the center navigation view. Some qualitative results are presented in Figure \ref{result_turn}.
	
	Figure \ref{result_turn}(a) presents the cases of different turns under various navigation instructions. The proposed model can well incorporate the traversability with the navigation instruction which produce correct paths towards the goal on vision images and generate well-shaped navigation cost map in geometric world. It can be verified in the navigation cost map that the predicted free space towards the goal can consistently avoid the obstacle detected by the laser.  Figure \ref{result_turn}(b) specifically shows the outstanding performance of dynamic obstacle recognition, including the driving cars (first column), the pedestrians (second column) and the moving bicycle (third column). Without the accurate pixel-level annotation of moving objects, the model has implicitly learned some driving common sense from demonstrated human drives. 
	
	\subsection{Multi-modal behavior study}
\begin{figure}[!h]
		\centering
		\includegraphics[width=0.48\textwidth]{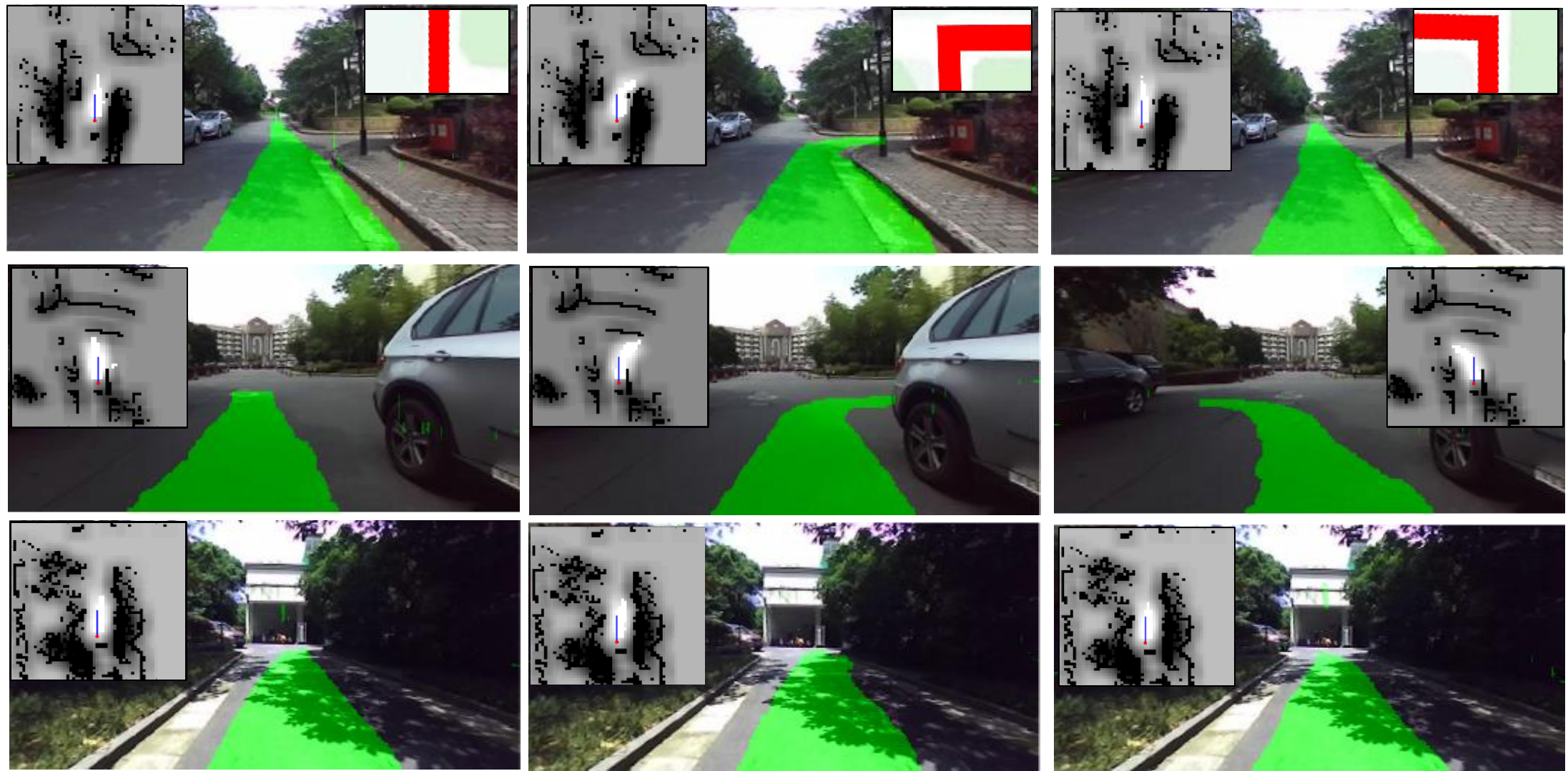}
		\caption{Multi-modal driving behaviors under different fake navigation instructions. From top to bottom: T-juction, crossroad and straight road.}
		\label{simulation}
	\end{figure}

	The weakly-supervised label generation can only validates the goal-directed path with one deterministic navigation instruction for each perception, while multi-modal behaviors are present at intersections and when approaching open areas. For a further qualitative evaluation, we have made three fake navigation instructions intuitively viewed as \{go-straight, turn-right, turn-left\} respectively to generate goal-directed paths for the same perceptions. Some typical results for different road sections have presented in Figure \ref{simulation}. 
	
\begin{figure}[!h]
		\centering
		\includegraphics[width=0.48\textwidth]{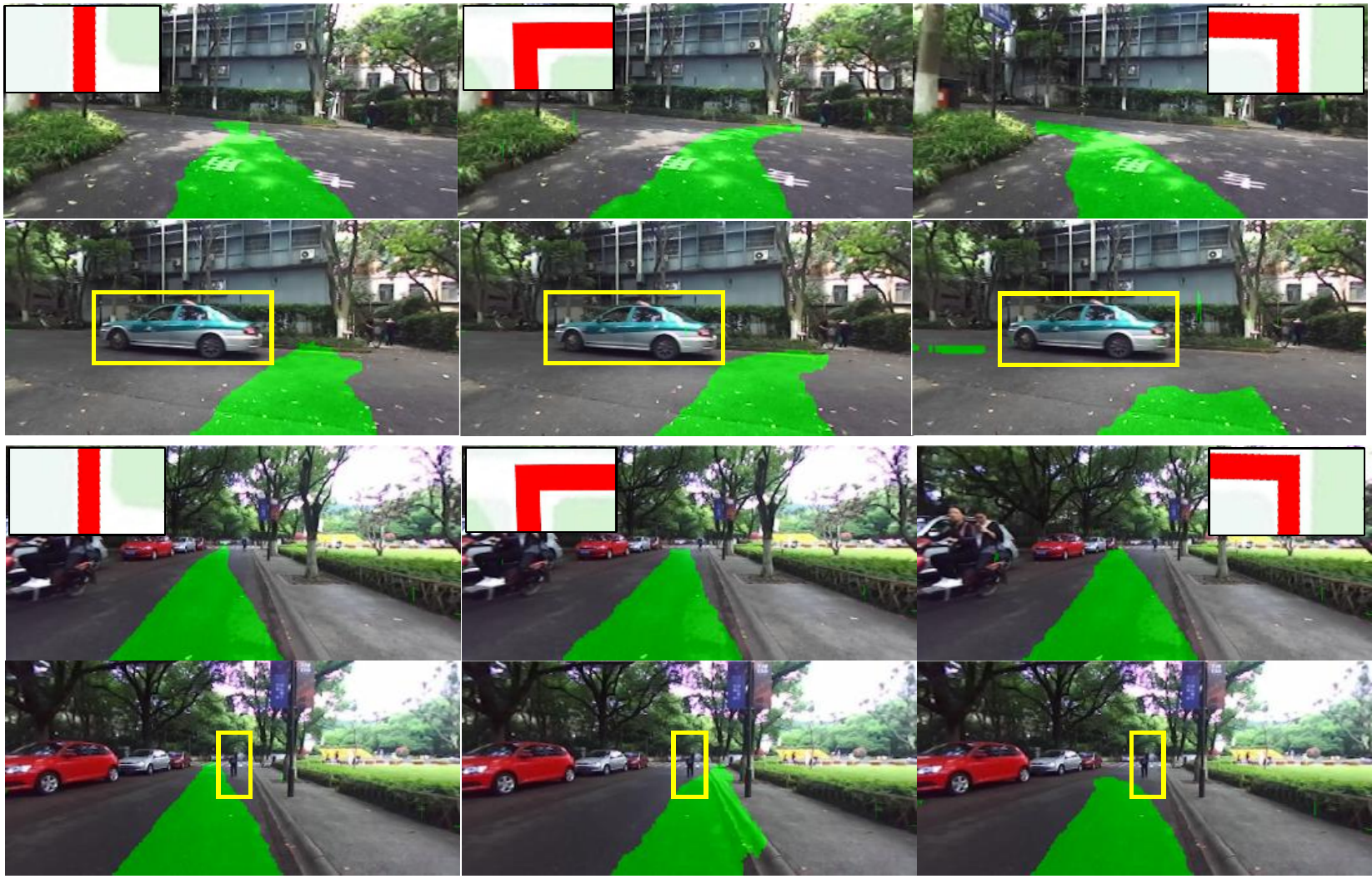}
		\caption{Multi-modal behavior comparison of path generation with/without dynamic objects. The top two rows and the bottom two rows respectively show the performance around one same place.}
		\label{simulation2}
	\end{figure}

	The three rows show different road types of T-juction, crossroad and straight road respectively. When the turning actions are not allowed in the given environment, the inherently learned traversability property of the model can still generate reasonable paths based on the road textures. Nevertheless, when there are obstacles appeared in the critical area along the road, it can have significant impact on path generation result, as shown in Figure \ref{simulation2}, where two groups of experiments compared under different navigation instructions with/without obstacles.
	
	\begin{table*}[!t]
		\caption{R{\scriptsize OBUSTNESS} {\scriptsize TO} L{\scriptsize OCALIZATION}  E{\scriptsize RRORS}}
		\centering
		\begin{tabular}{lccccccccc}
			\toprule[0.025cm]
			\multicolumn{1}{l}{\multirow{2}{*}{}} &
			\multicolumn{3}{c}{minor (0$\sim$6\%)}&\multicolumn{3}{c}{moderate (6\%$\sim$12\%)}&\multicolumn{3}{c}{hard (12\%$\sim$18\%)}\\
			\cmidrule(lr){2-4} \cmidrule(lr){5-7} \cmidrule(lr){8-10}
			Algorithm &IOU/\% & $cover\_rate$/\%&$\Delta{yaw}/deg$&IOU/\% & $cover\_rate$/\%&$\Delta{yaw}/deg$&IOU/\% & $cover\_rate$/\%&$\Delta{yaw}/deg$\\
			\cmidrule{1-10}
			\textit{basic} & 58.81&99.08&9.26 & 57.78&99.11&9.86& 56.10&98.48& 11.52  \\
			\cmidrule{1-10}	
			\textit{$rand\_navi$} &61.48& 98.83 &10.34& 64.90&99.22&10.27& 65.65&99.43& 9.72\\
			\cmidrule{1-10}
			\textit{$rand\_per$}  &55.20&96.35&8.82&54.74&95.82&8.91&55.79&96.23&9.42\\
			\bottomrule[0.025cm]
		\end{tabular}
		\label{robustness to localization}
	\end{table*}
	
	The upper two rows show the comparison results for a moving car. The model outputs directional path specifically related to different instructions when the front space is free. However when there is a car appeared in the left part, the \textit{go-straight} instruction has been influenced by a right shift and the \textit{turn-left} has failed to generate a well-shaped path area. In the lower group of images (3-4 rows), shows a straight road section. When there is a pedestrian approaching, the paths generated show a slight deviation to accommodate the moving obstacle. The generated path has covered a bit to the curbs in the middle figure. This is caused by the inaccurate annotation in the training data which has also traversed to some curb areas, and can be avoided by a narrower assumption of vehicle width.

	\subsection{Robustness to Localization errors}
	
	This section discusses the model prediction robustness of the path generation with potential localization errors. In order to achieve it, we have randomly added offsets on the navigation instructions. This has resulted in deviation of the centre line of the annotated route with local navigation view. 
	This can be efficiently implemented using the method illustrated in Section\ref{labelgeneration}. The random offset goes in three level strengths, easy, moderate and hard, with each level corresponding to offset of $0\sim6\%$, $6\sim12\%$, and $12\sim18\%$ the image size. This approximately represents $\sim 1 m$, $\sim 2.5m$,$\sim 4m$ geometry offset from the route center. The performance of different models is presented in Table \ref{robustness to localization}
	
\begin{figure}[!h]
		\centering
		\includegraphics[width=0.45\textwidth]{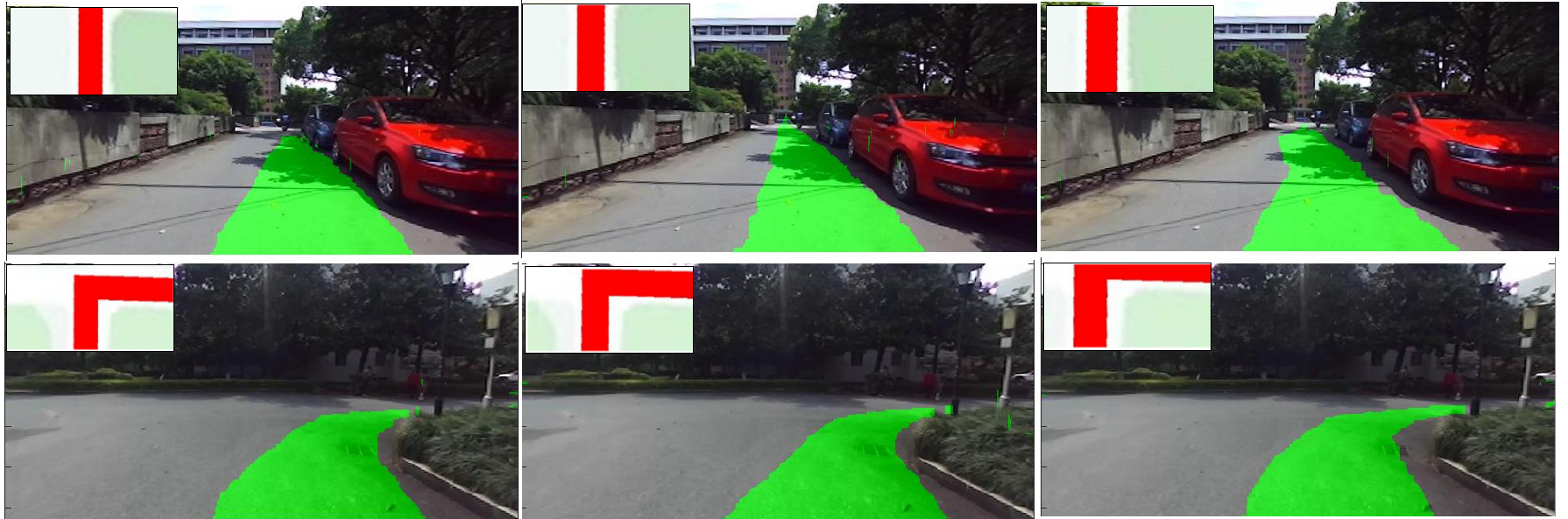}
		\caption{Visual result of the robustness to localization.}
		\label{localization}
	\end{figure}

	As can be seen in the table, $cover\_rates$ of different models have remain in a similar value to the previous center view navigation result, indicating a consistent main direction to the provided real mask. As expected, the $basic$ model performance was affected by harsher navigation offsets in terms of IOU and yaw angle. This is apparently visible in the images.  
	In contrast, the $rand\_navi$ behaves slightly better when the offsets were introduced. This may be due to utilization of varied navigation instructions for a certain perception in the training phase. However,the training setting of fixed perception/mask with varied navigation instructions can negatively influence the generated path.  
	Since the direction distinctions always take a small share on the image, it can lead to high $iou$ but low geometric angle accuracies. The $rand\_per$ model remains relatively unaffected with a relatively slight drop of $\Delta{yaw}/deg$. The random offset on perception has once again let the model pay attention to the road structure which enables this model behaves most robust to localization errors. Some qualitative results from the $rand\_per$ with different instruction offsets are provided in Figure \ref{localization}.
	
	\subsection{Prediction ability of potential maps}
	In this section, we study the prediction ability of potential maps using the ground truth trajectory of test routes obtained by the traditional localization algorithm \cite{tang2017one}. The trajectory within a maximum distance in front of current position is considered for performance evaluation. The grids where robot pose located receive both the positive plausibility from the free space and the negative plausibility from the obstacle. 

	Thus the NLL cost can be calculated using the trajectory length, the result is shown in Table \ref{geometric evaluation}.
	
	\begin{table}[!h]
		\caption{T{\scriptsize RAJECTORY} C{\scriptsize OST} E{\scriptsize STIMATION}}
		\centering
		\begin{tabular}{lccc}
			\toprule[0.025cm]
			Prediction distance &$basic$&$rand\_navi$&$rand\_per$\\
			\cmidrule{1-4}	
			\textit{5 m} & 0.178&0.208&0.192\\
			\cmidrule{1-4}	
			\textit{10 m} & 0.476&0.514&0.384\\
			\bottomrule[0.025cm]
		\end{tabular}
		\label{geometric evaluation}
	\end{table}
	
\begin{figure}[!h]
		\centering
		\includegraphics[width=0.47\textwidth]{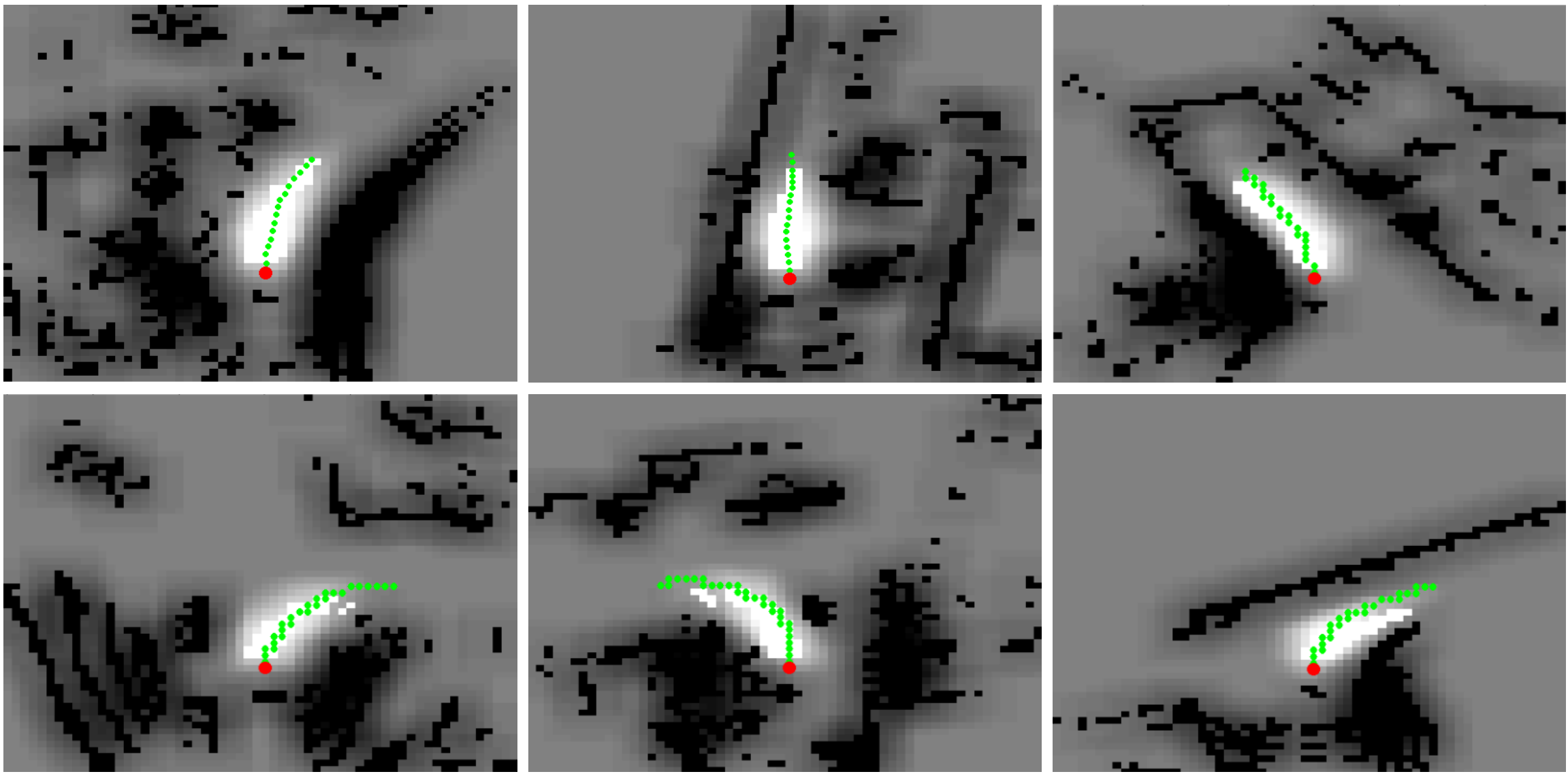}
		\caption{Trajectory prediction using generated potential map. The red dot  shows the location of the robot where as the green dots represent the ground truth trajectory of one human driver.}
		\label{potentialmaps}
	\end{figure}

	Intuitively, less prediction distances result in superior performances. And the probability of human driver in the cost map is relatively 86\% for the 5m distance and 68\% for the 10m distance. Similarly, $basic$ behaves best in the simple case and $rand\_per$ behaves better in more complicated settings. Some visually result from $rand\_per$ is shown in Figure \ref{potentialmaps}, the green points represent the robot trajectory. 
	
\begin{figure}[!h]
		\centering
		\includegraphics[width=0.43\textwidth]{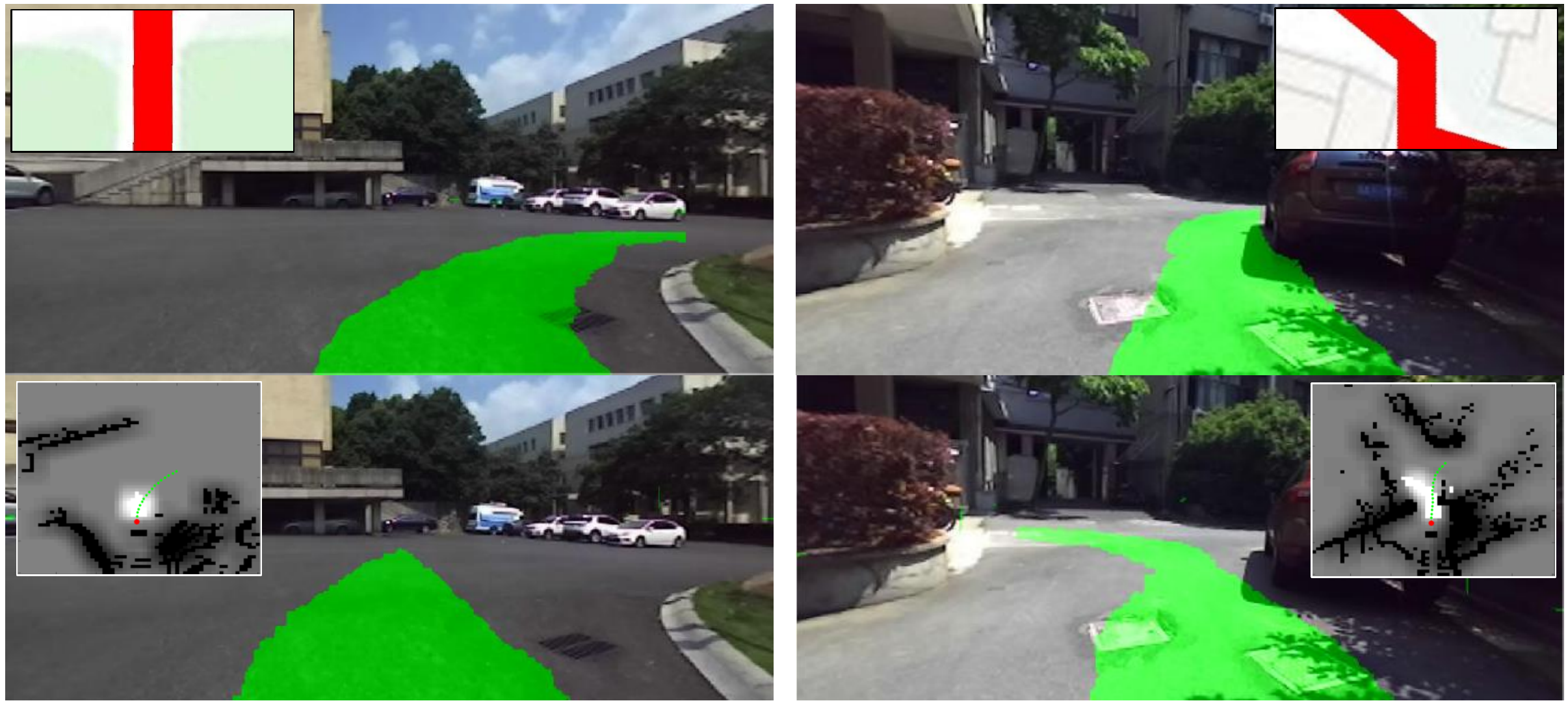}
		\caption{Example of the prediction failures. The top images are the ground truth, and the bottom images are their predictions with the cost map shown in the corner. }
		\label{failure}
	\end{figure}

	The first row shows the examples with a trajectory probability exceeds 0.85, and most of the poses lie in the central of low cost area. The second row shows some examples with lower confidence visually correct results. The main reason of the lower confidence goes into two sides. First is the limited prediction distance near turns with monocular camera, and the second is the difference on specific turning curvatures. Some of the incorrect predictions are shown in Figure \ref{failure}.
	
	The upper row is the ground truth while the lower is the prediction. This poor performance is attributed to significantly large offsets introduced between the current position to that of the navigation route. In the case of the left figure, a delayed navigation signal was introduced while for the right image, an advanced navigation signal was introduced.  
	Although some of the failures have provided deviated path directions from the ground truth, they are basically located in the safe area and have the potential to fix its direction with the new timely instruction. In conclusion, both of the quantitative and qualitative results have implied the navigation cost map can be used for the motion planning to mimic the expected driving behaviors with a human driver. 
	
	%
	%
	%

	\section{CONCLUSIONS}\label{conclusion}
	In this paper, we have proposed an innovative navigation framework that learns to cast current robot perception along with the GPS navigation instruction directly to the geometric navigation cost map. The system has adopted a deterministic conditional adversarial network to encourage the path diversity under the constrain of road structure and specified instruction. It learns a model in a weakly supervised manner incorporating road visual semantics. 
	The method has achieved multi-modal driving behavior prediction in varied scenarios with  specified navigation instructions. The prediction ability of the navigation cost maps were experimentally validated with the real human driver trajectories, which can be efficiently utilized for robot motion planning and control command generation. Our future work will consider the learning of specific driving rules and cognition of advanced common sense for a better flexibility and safety.

	\section*{ACKNOWLEDGMENT}
	This work was supported in part by the National Key R\&D Program of China (2017YFB1300400), in part by the National Nature Science Foundation of China (U1609210).
	\medskip 
	\bibliographystyle{IEEEtran}
	\bibliography{mybib}
	
\end{document}